\documentclass[10pt,twocolumn,letterpaper]{article}
\usepackage[pagenumbers]{cvpr} 

\usepackage[table, dvipsnames]{xcolor}

\definecolor{cvprblue}{rgb}{0.21,0.49,0.74}

\usepackage[pagebackref,breaklinks,colorlinks,citecolor=cvprblue]{hyperref}

\usepackage{multirow}
\usepackage{verbatim}
\usepackage{pifont}
\newcommand{\cmark}{\ding{51}}

\newcommand{\pub}[1]{{\color{gray}{\scriptsize{[{#1}]}}}}
\usepackage{booktabs}
\usepackage{arydshln} 
\definecolor{mygray}{gray}{.9}
\definecolor{mygray1}{gray}{.97}

\makeatletter
\newcommand{\thickhline}{
    \noalign {\ifnum 0=`}\fi \hrule height 1pt
    \futurelet \reserved@a \@xhline
}



\def\paperID{2699} 
\def\confName{CVPR}
\def\confYear{2024}

\title{MGMap: Mask-Guided Learning for Online Vectorized HD Map Construction}

\author{Xiaolu Liu$^1$,  \ \ Song Wang$^1$,  \ \ Wentong Li$^1$,  \ \  Ruizi Yang$^1$,   \ \   Junbo Chen$^{2}\footnotemark[1]$, \ \ \ Jianke Zhu$^{1}\thanks{Corresponding authors}$\\
	$^1$Zhejiang University \ \ \
	$^2$Udeer.ai \\ 
	{\tt\small \{xiaoluliu, songw, liwentong, ruiziyang, jkzhu\}@zju.edu.cn, junbo@udeer.ai}
}

\begin{document}
\maketitle

\begin{abstract}
Currently, high-definition (HD) map construction leans towards a lightweight online generation tendency, which aims to preserve timely and reliable road scene information.
However, map elements contain strong shape priors. Subtle and sparse annotations make current detection-based frameworks ambiguous in locating relevant feature scopes and cause the loss of detailed structures in prediction. To alleviate these problems, we propose MGMap, a mask-guided approach that effectively highlights the informative regions and achieves precise map element localization by introducing the learned masks. Specifically, MGMap employs learned masks based on the enhanced multi-scale BEV features from two perspectives. At the instance level, we propose the Mask-activated instance (MAI) decoder, which incorporates global instance and structural information into instance queries by the activation of instance masks. At the point level, a novel position-guided mask patch refinement (PG-MPR) module is designed to refine point locations from a finer-grained perspective, enabling the extraction of point-specific patch information. Compared to the baselines, our proposed MGMap achieves a notable improvement of around 10 mAP for different input modalities. Extensive experiments also demonstrate that our approach showcases strong robustness and generalization capabilities. Our code can be found at \href{https://github.com/xiaolul2/MGMap}{https://github.com/xiaolul2/MGMap}.
\end{abstract}
\vspace{-3mm}
\section{Introduction}
\label{intro_fig}
\begin{figure}
  \centering
  \includegraphics[width=0.45\textwidth]{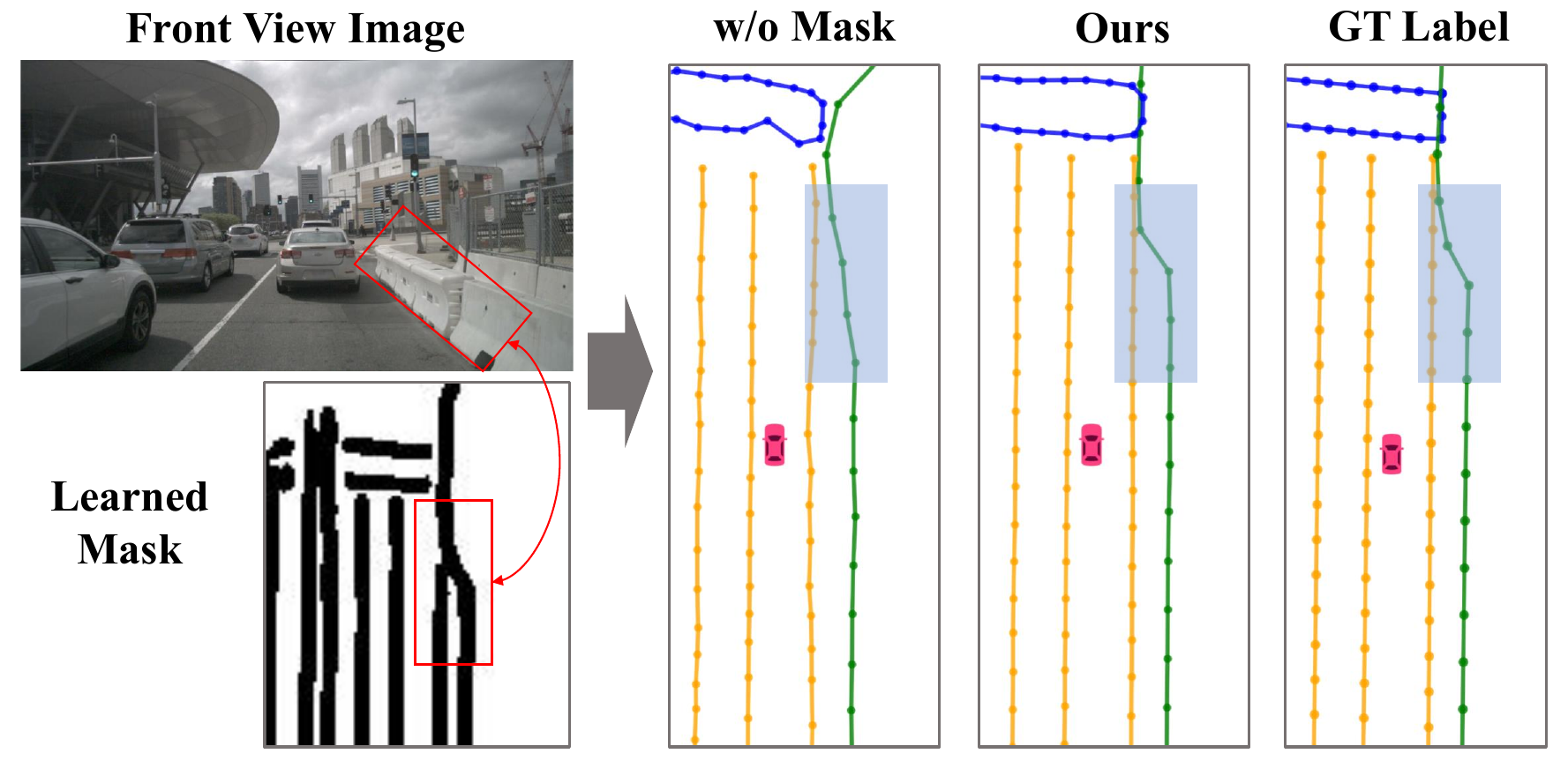}
  \vspace{-3mm}
  \caption{For some detailed structures, our proposed MGMap achieves effective map element localization by highlighting the informative regions through the learned masks.}
  \vspace{-4mm}
  \label{fig0}
  \vspace{-2mm}
\end{figure}
High-definition (HD) map plays an important role in autonomous driving~\cite{li2022hdmapnet}, as it provides centimeter-level road information for self-positioning~\cite{levinson2007map}, path planning~\cite{gao2020vectornet,da2022path} and other downstream tasks~\cite{hu2023planning,he2023egovm,jiang2022perceive}. Typically, offline HD map constructions rely on manual annotations on global LiDAR point clouds,
which usually require tedious financial and manual investments. Moreover, it is challenging for offline maps to reflect the ever-changing road conditions due to the lack of real-time updates. To tackle the above issues, a lightweight and real-time online map construction paradigm~\cite{li2022hdmapnet} has gradually become a promising approach by incorporating information from onboard sensors.

Some existing approaches~\cite{roddick2020predicting,li2022hdmapnet,zhang2022beverse,dong2022superfusion,wang2023lidar2map} consider online map construction as a segmentation task, where pixel-level rasterized maps are learned in bird's-eye-view (BEV) space. 
Nevertheless, for vectorized expression, an additional step of post-processing is required to cluster and fit lane instances. 
Recently, efficient and straightforward approaches like VectorMapNet~\cite{liu2023vectormapnet} and MapTR~\cite{liao2022maptr} have been proposed for vectorized map construction, in which map elements are represented by sparse point sets. Transformer-based architectures are directly employed to update instance queries and regress point locations. 

Despite having achieved promising results, current map vectorization frameworks are still constrained by inherent issues. As shown in~\Cref{fig0}, map elements, such as road edges, dividing lines, and pedestrian crossings, always have strong shape priors. Obscure features and coarse locations can easily lead to the loss of detailed expressions in prediction, especially for irregular boundaries and sudden changes in corner angles. Besides, subtle and sparse annotations pose significant challenges for deformable attention~\cite{zhu2020deformable}, which is known as a sparse and local feature extraction strategy employed in current vectorization frameworks. Considering the sparsity of detection targets, such under-sampling techniques can readily lead to coarse localization and the loss of effective information. These issues intensify the difficulties in determining the relevant feature scopes, confirming lane line instances, and accurately locating lane points. 

To tackle the above issues, in this paper, we propose a fine-grained approach called MGMap, which aims to improve localization and highlight specific features by incorporating the guidance of learned map masks. Initially, the enhanced multi-level BEV features are dynamically constructed for richer semantic and position features. Based on this, masks can be generated and adequately utilized at the holistic instance level and the more granular point level. At the instance level, we design the Mask-Activated Instance (MAI) decoder to guide the construction and feature aggregation of lane queries. By leveraging learned instance masks, MAI decoder enables the activation of lane queries to possess global instance structures and shape characteristics upon activation. Furthermore, at the point level, a Position-Guided Mask Patch Refinement (PG-MPR) module is proposed to alleviate the difficulty in locating relevant features due to the sparse and irregular detection targets. By focusing on the specific patch regions, binary mask features can be extracted to gather more detailed information at lanes' surrounding locations, which is well-designed for finer regression of the detailed structure and point locations. 

Extensive experiments on \textit{nuScenes}~\cite{nuScenes} and \textit{Argoverse2}~\cite{Argoverse2} datasets demonstrate that MGMap achieves state-of-the-art (SOTA) performance on the task of online HD map construction. Besides, the promising experiment results under different settings show the robustness and generalization capability of our presented model. The main contributions of our work can be summarized as follows:
\begin{itemize}
    \item An effective approach for precise online HD map vectorization with the guidance of learned masks. The effective features of instance masks and binary masks are extracted for unique lane lines and shape learning.
    \item A mask-activated instance decoder and a novel position-guided mask patch refinement module to decode map elements from the instance level and the point level by fully leveraging the potential of mask features. 
    \item Promising results on two testbeds show that our MGMap outperforms the previous approaches at a large margin and has strong robustness and generalization capability.
\end{itemize}

\begin{figure*}[t]
\begin{center}
\centering
\includegraphics[width=1.0\textwidth]{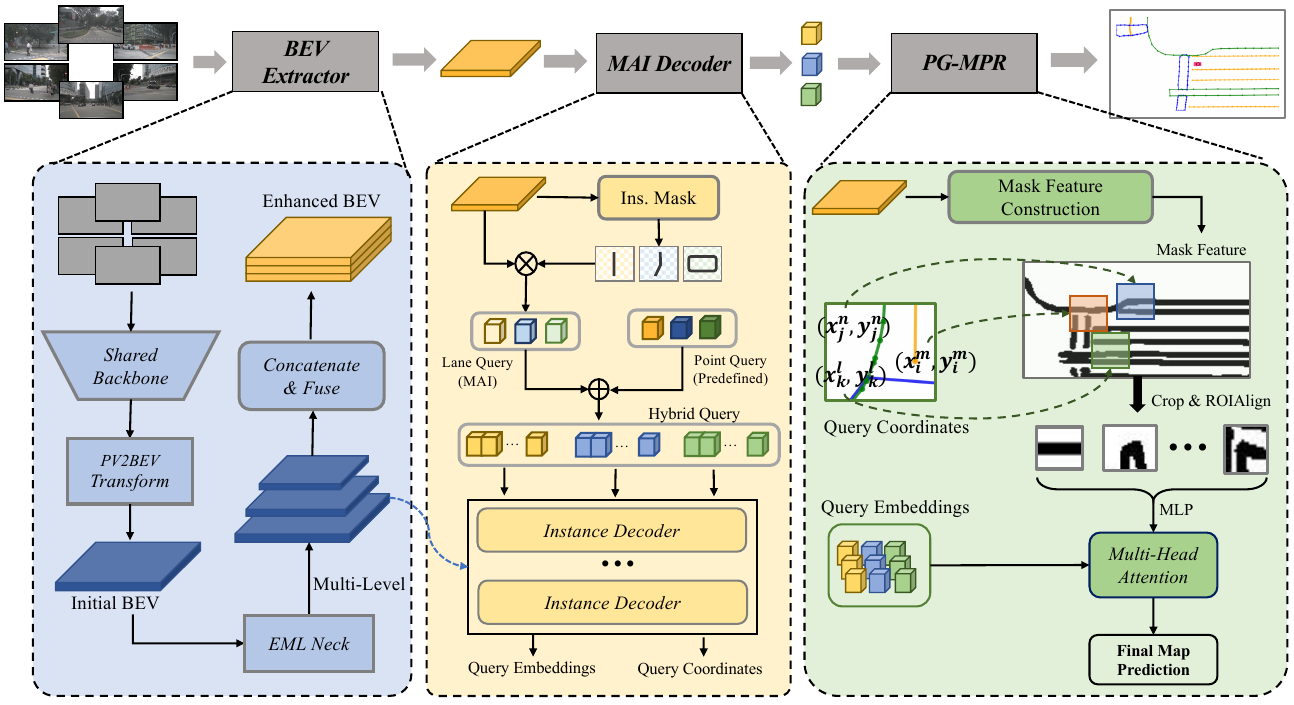}
\vspace{-6mm}
\caption{\textbf{Overview of MGMap framework.} MGMap mainly consists of three components: (1) BEV Extractor to obtain multi-scale BEV features by transforming from perspective view (PV) to BEV with the enhanced multi-level neck; (2) Mask-Activated Instance (MAI) Decoder is employed to construct and update queries at instance level; (3) Position-Guided Mask Patch Refinement (PG-MPR) module is designed to refine points' positions from local patch features at point level.}
\label{fig_framework}
\vspace{-6mm}
\end{center}
\end{figure*} 
\section{Related Work}
\label{sec:relatedwork}
\noindent\textbf{Online HDMap Construction.}
Unlike traditional offline HD map annotations on global LiDAR points~\cite{shan2018lego,shan2020lio}, recent studies~\cite{li2022hdmapnet, wang2023lidar2map,liu2023vectormapnet,liao2022maptr} explore the online construction directly from on-board sensor data to reduce the cost of labeling and provide up-to-date road information. Some approaches~\cite{li2022hdmapnet,li2022bevformer,wang2023lidar2map,dong2022superfusion,qin2023unifusion} treat HD map construction as a segmentation task to predict pixel-level rasterized maps, which requires post-processing for vectorized construction. For more straightforward vectorized construction, Liu~\etal \cite{liu2023vectormapnet} propose a two-stage framework VectorMapNet with an auto-regressive decoder to recurrently connect vertices. Jeong~\etal~\cite{shin2023instagram} detect points first and employ an adjacent matrix in InstaGraM to build the connections among instance points. 
Further, Liao~\etal \cite{liao2022maptr, liao2023maptrv2} propose to represent map elements as the ordered points with fixed numbers in MapTR, so that transformer architecture can be used to regress points' positions simultaneously. Later, Qiao \etal\cite{qiao2023end} and Ding \etal\cite{ding2023pivotnet} present new modeling strategies, in which BeMapNet and PivotMap utilize Bezier curves and dynamic pivot points to model map elements separately for more detailed representations. In contrast to the above methods, our MGMap approach introduces the guidance of generated mask features to handle lane shapes in detail with specific feature enhancement.

\noindent \textbf{Camera-based BEV Perception.} HD map construction relies on high-quality BEV features, which are also the basis for most 3D perception tasks~\cite{wang2021fcos3d, huang2021bevdet, liu2022petr, philion2020lift, wang2022detr3d}. Generally, BEV features are extracted and transformed from perspective-view (PV) images. Initially, Reiher~\etal \cite{reiher2020sim2real} utilize homography transformation to project PV images into BEV space in IPM. After that, learning-based approaches are widely used to construct more reliable BEV features~\cite{can2021structured,wang2021fcos3d}. Pan~\etal \cite{pan2020cross} employ a fully connected layer in VPN to convert perspective view features into BEV space. In~\cite{philion2020lift} and~\cite{huang2021bevdet}, depth estimation is utilized to establish the connection between the surrounding view images and BEV features. In order to enhance the robustness of the model and the effect of detection, multi-modality and temporal fusion strategies are used in BEVFormer \cite{li2022bevformer} and BEVFusion \cite{liu2023bevfusion, liang2022bevfusion}. Besides, transformer-based architectures~\cite{peng2023bevsegformer,liu2022petr,li2022bevformer} are widely used for feature aggregation. BEV features are represented by queries at different positions, which can be updated by interaction with PV image features. To obtain a more reliable HD map, we employ a pyramid-like network for multi-scale BEV features, which enables to capture the rich semantic and location information for online HD map construction.

\noindent\textbf{Mask Refinement for Segmentations.} 
Mask refinement strategies are widely used for different segmentation tasks, aiming to improve the quality of instance or semantic features.  In previous work~\cite{cheng2020boundary} and~\cite{takikawa2019gated}, boundary-aware mask features are constructed by an extra branch, which is used to improve mask localization accuracy. Cheng~\etal~\cite{cheng2022sparse} utilize the interaction between learned mask features with instance activation maps for representing objects with instance features. In~\cite{liang2020polytransform,cheng2021per, cheng2022masked, ding2023hgformer}, mask features are integrated into the transformer-based architecture, in which attentions are utilized for feature extraction. Based on the initial prediction outputs, Tang~\etal~\cite{tang2021look} propose a refinement strategy with small boundary patches to improve the mask quality. As for map construction, our proposed MGMap approach takes advantage of the learned mask features from both the instance level and point level to highlight and enhance informative regions of subtle map annotations.

\section{MGMap}
\label{sec:method}
Given surround-view images captured from onboard cameras, our goal is to distinguish local BEV map instances while locating their corresponding structures. Each map element comprises a class label $\mathbf{c}$ and an ordered sequence of points $\mathbf{P}=\{(x_i,y_i)\}_{i=1}^{N}$ that represents the lane structure, with $N$ denoting the number of points for each lane.

 \Cref{fig_framework} illustrates the framework of our proposed MGMap. To obtain multi-scale BEV features, a pyramid network with fused attention is designed, which is built after initial PV-to-BEV feature transformations (\Cref{part1}). Subsequently, a cascaded transformer decoder is utilized to update mask-activated instance queries (\Cref{part2}). Finally, by interacting with the local semantic context derived from the patch of mask features, we achieve fine-grained position refinement at the point level (\Cref{part3}).

\subsection{BEV Feature Extraction}
\label{part1}
Initially, we employ shared CNN backbones to extract 2D features from PV images. PV features can be gathered into a unified BEV representation via the strategy in ~\cite{li2022bevformer}, which utilizes deformable attention~\cite{zhu2020deformable} to update BEV queries by interacting with surround-view image features. The extracted BEV feature can be represented as $\mathbf{F}_0 \in \mathbb{R}^{D \times H_{BEV} \times W_{BEV}}$, where $H_{BEV} \times W_{BEV}$ is the size of the BEV feature and $D$ represents the dimension.

\noindent\textbf{Enhanced Multi-Level Neck.} 
To obtain BEV features with rich semantic and location information, 
at BEV space, we design a 3-layer Enhanced Multi-Level (EML) neck with fused attention to construct the unified BEV features. Multi-scale BEV features with larger receptive fields can be obtained for a better understanding of the overall structures. 

In the EML neck, we construct the cascaded layers using residual blocks based on the hybrid of channel attention (CA) and spatial attention (SA)~\cite{woo2018cbam}, in which hybrid multiplications among channels and spaces are designed to capture both local and global contextual information. Such dynamic selections at different BEV spaces are beneficial for detecting lane lines with irregular shapes. The calculation of learnable attention maps can be formulated below
\begin{equation}
    \mathbf{F}_{i+1} = (\text{CA}(\mathbf{F}_{i}) \times 
    \mathbf{F}_{i}) \times \text{SA}(\mathbf{F}_{i}).
\end{equation}
After that, multi-level BEV features $\{\mathbf{F}_i\}_{i=1}^3$ can be obtained with the shape of ${D}_i\times\frac{H_{BEV}}{2^{i+1}}\times\frac{W_{BEV}}{2^{i+1}}$ for each layer $i$, where $D_i$ is the corresponding dimension. To preserve all the information at different levels, we employ bilinear interpolation to upsample 3-level outputs $\{\mathbf{F}_i\}_{i=1}^3$. All these features are aligned to have the same resolution as the initial $\mathbf{F}_0$. Finally, a $3 \times 3$ convolutional layer after concatenation is used to aggregate multi-level features and obtain the enhanced BEV features $F_c$. Such enhanced $F_c$ contains local and semantic information across varying receptive fields, enhancing the capability to detect fine-grained structures.

\subsection{Mask-Activated Instance Decoder}
\label{part2}
For each lane instance, specific query embeddings with instance and structure information are required for the regression of lane shapes and positions. Based on enriched BEV features, this section focuses on the design of mask-activated lane queries and the subsequent update through the cascaded deformable transformer decoder.

\noindent \textbf{Mask-Activated Query.}
To achieve a more detailed and specific representation, MGMap employs a hybrid approach that combines lane queries $\mathbf{Q}_{lane}$ and point queries $\mathbf{Q}_{point}$ to encode individual map instances.

\begin{figure}[t]
  \centering
  \includegraphics[width=0.48\textwidth]{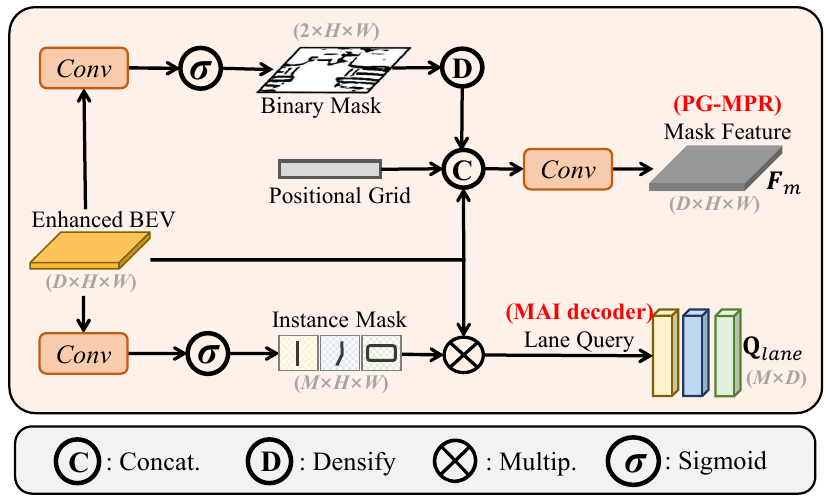}
  \vspace{-6mm}
  \caption{Illustration of \textbf{mask constructions} at different stages. In MAI decoder, instance masks are generated to activate lane queries, while binary masks are extracted to provide fine-grained patch features in PG-MPR.}
  \vspace{-5mm}
  \label{fig_Mask}
\end{figure}

For lane queries, in contrast to the fixed query designs~\cite{liao2022maptr,zhu2020deformable}, $\mathbf{Q}_{lane}\in\mathbb{R}^{M \times D}$ is dynamically initialized with the guidance of learned instance masks, which is more flexible and contains specific shape prior and instance information. Here $M$ is the number of instance queries and $D$ is the dimension. Specifically, we obtain a set of instance segmentation maps $\mathbf{M}_{ins}\in\mathbb{R}^{M \times H_{BEV}\times W_{BEV}}$ by applying basic convolution with the sigmoid function $\sigma(\cdot)$ to the enhanced BEV feature $\mathbf{F}_c$, as illustrated in the lower half of~\Cref{fig_Mask}. By leveraging $\mathbf{M}_{ins}$, we generate the mask-activated instance query $\mathbf{Q}_{lane}$ via multiplication between instance masks $M_{ins}$ and the transpose of enhanced BEV features
\begin{equation}
    \mathbf{Q}_{lane}=(\sigma((\text{Conv}(\mathbf{F}_c))) \times \mathbf{F}_c^\top.
\end{equation}
During the training stage, $\mathbf{M}_{ins}$ is supervised by instance mask annotations, and bipartite matching is used to pair each instance mask with the ground-truth label. This activated instance query allows for a more tailored and precise representation of each map element, in which unique features in instance masks can be aggregated to the specific instance queries.

For point queries, $\mathbf{Q}_{point}\in\mathbb{R}^{N \times D}$ is constructed using a set of predefined learnable weights, which are broadcasted by multi-layer perceptions (MLP) to fuse with each $\mathbf{Q}_{lane}$.
Thus, the hybrid query $\mathbf{Q}\in \mathbb{R}^{M\times N\times D}$ can be obtained by
    \begin{equation}
     \mathbf{Q}=\mathbf{Q}_{lane} + \text{MLP}(\mathbf{Q}_{point}).
    \end{equation}

\noindent \textbf{Deformable Decoder.} To update the hybrid quires, an $L$-layer decoder can be built by following the multi-scale deformable DETR~\cite{zhu2020deformable} structure. At each layer $l$, query embeddings can be updated by interacting with structured multi-level BEV features. Specifically, deformable DETR assigns reference points $\mathbf{P}^l$ as anchors to collect sparse features from $\{\{\mathbf{F}_i\}_{i=1}^3, \mathbf{F}_c\}$, in which $\mathbf{P}^l$ is also the intermediate stage of the normalized lane point positions. Simultaneously, the updated lane point positions $\mathbf{P}^{l+1}$ can be obtained by adding the learned offsets, which are derived from MLP regression branch $\text{Reg}_{pos}$ on query $\mathbf{Q}^l$ as follows
\begin{gather}
    \mathbf{Q}^l=\text{DeformAttn}(\mathbf{Q}^{l-1}, \mathbf{P}^l,\{\mathbf{F}_i\}_{i=1}^3,\mathbf{F}_c), \\
    \mathbf{P}^{l+1} = \sigma(\sigma^{-1}(\mathbf{P}^l) +\text{Reg}_{pos}(\mathbf{Q}^l)).
\end{gather}
The classification scores $\mathbf{c}^{l+1}$ can also be updated by MLP regression on $\mathbf{P}^l$.
With iterative interaction in the $L$-layer decoder, MGMap can progressively update the semantic and positional information to query embeddings and constantly revise the shape and position of lane lines.

\subsection{Position-Guided Mask Patch Refinement}
\label{part3}
Although the general shapes and structures can be regressed from the instance level, some detailed expressions are still hard to construct. Therefore, it is necessary to achieve fine-grained refinement at a more specific point level. 
In this section, we design a refinement module to utilize binary mask features more specifically. 

\noindent \textbf{Mask Feature Construction.}
As shown in \Cref{fig_Mask}, binary mask $\mathbf{M}_{b} \in \mathbb{R}^ {2\times H_{BEV} \times W_{BEV}}$ can be obtained by applying sigmoid function after basic convolution on $\mathbf{F}_c$, in which auxiliary loss with rasterized supervision is used at the training stage. With the guidance of binary segmentation learning, more highlighted features can be utilized to distinguish lane line information from the background. 

 After that, the binary mask feature $\mathbf{F}_{m}$ can be constructed based on the binary map $\mathbf{M}_{b}$. We densify the dimension of $\mathbf{M}_{b}$ from 2 to 32 by $\text{D}(\cdot)$. Then, we concatenate the densified binary map with $\mathbf{F}_c$ and a 2-channel normalized positional grid $\mathbf{G}_{bev}$, which comprises spatial local information features for each pixel to provide relative location information. Finally, these three kinds of characteristic features are fused by the convolutional operation as below
\begin{equation}\label{b}
    \mathbf{F}_m = \text{Conv}(\text{Concat}(\mathbf{F}_{c}, \text{D}(\mathbf{M}_{b}),\mathbf{G}_{bev})).
\end{equation}
Compared to $\mathbf{F}_c$, $\mathbf{F}_{m}$ emphasizes specific location and semantic information around lane lines to differentiate the feature from ambiguous backgrounds, which provides more clear location information. 

\noindent \textbf{Patch Extraction and Refinement.}
\begin{figure}
  \centering
  \includegraphics[width=0.47\textwidth]{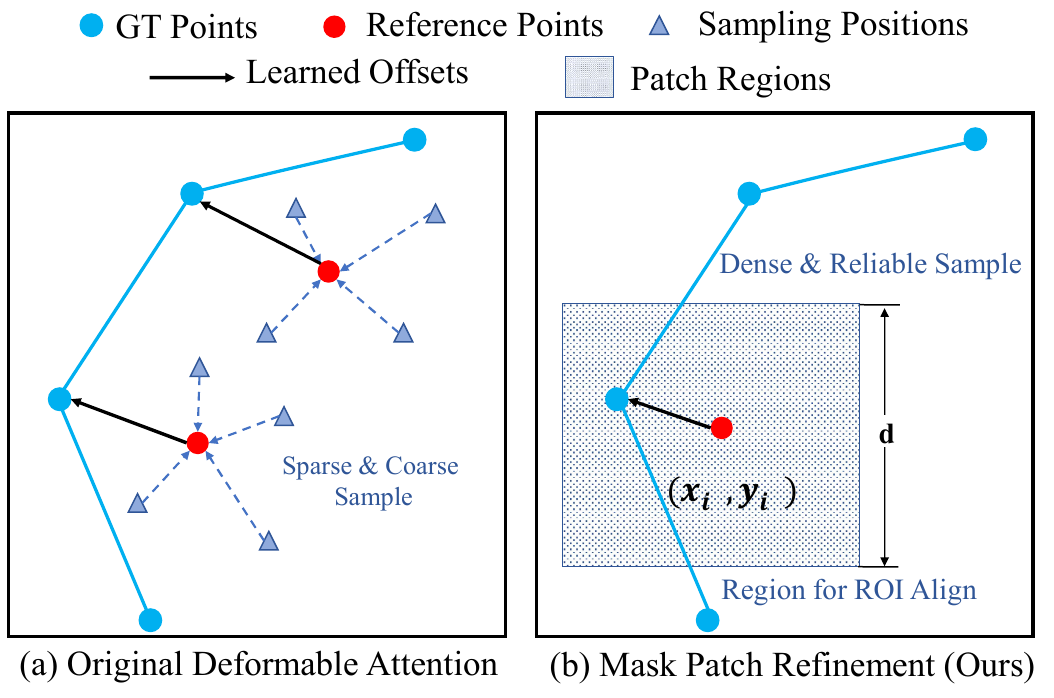}
  \vspace{-2mm}
  \caption{(a) The conventional deformable attention extracts sparse features from sampling points, which may select irrelevant features; (b) Our proposed \textbf{Mask Patch Refinement} extracts more relevant features from the region of reliable patch.
  }
  \label{fig4}
  \vspace{-4mm}
\end{figure}
As shown in \Cref{fig4} (b), for the $i$-th point of the lane line, the patch region is a squared bounding box $B_i$ centered at the point's coordinates $(x_i,y_i)$, which is regressed from the last layer of the MAI decoder. The size of the patch region is determined by the hyper-parameter $d$.

Once the location and region scale are determined, we apply the function $f_{ext}$, to extract local patch features $\mathbf{v}_i$ for each point in each map instance. This process is implemented using ROIAlign \cite{he2017mask}, which is employed to extract and unify the features as follows
\begin{equation}
    \mathbf{v}_i= f_{ext}\big(\mathbf{F}_{m}, \text{R}( B_i ), (x_i,y_i)),
\end{equation}
 where $\text{R}(\cdot)$ denotes the denormalized patch region computation. The extraction function $f_{ext}$ utilizes basic convolutional blocks after bilinear interpolation and pooling to locate and align the corresponding values in the patch features. Consequently, each extracted semantic patch feature $\mathbf{v}_i$ has the shape of $D\times5\times5$, which effectively mitigates the issue of misalignment. Compared to the sparse deformable attention design in \Cref{fig4} (a), the mask patch design contains dense and more reliable features.

With query embedding $\mathbf{Q}$ from the L-layer MAI decoder and patch region feature $\mathbf{V}=\{\mathbf{v}_i\}_{i=0}^N$, we leverage multi-head attention to refresh the query features, which can be written as follows
\begin{equation}
    \mathbf{Q}^{s+1} = \text{MultiHeadAtt}\big( \frac{\mathbf{Q}^s \mathbf{V}^s}{\sqrt{D}}  \big),
\end{equation}
where $s$ represents the stage in the PG-MPR module and $D$ is the dimension of features. Query embeddings can be refreshed at a more detailed and specific point level. 

Finally, point coordinates and the classification scores can be regressed from query $\mathbf{Q}^s$ through MLP branches $\text{Reg}_{pos}$ and $\text{Reg}_{cls}$, which can be formulated as
\begin{gather}
    \mathbf{P}^{s+1} = \sigma(\sigma^{-1}(\mathbf{P}^s) +\text{Reg}_{pos}(\mathbf{Q}^s)), \\
     \mathbf{c}^{s+1} = \text{Reg}_{cls}(\mathbf{Q}^{s}).
\end{gather}

\subsection{Training Loss}
MGMap is trained in an end-to-end manner. Bipartite matching is employed to pair predicted map instances with their ground-truth counterparts. With the regression of points and class labels, auxiliary loss is required for mask segmentation. Concretely, the total loss is the sum of detection loss and mask segmentation loss $\mathcal{L} = \mathcal{L}_{det} + \mathcal{L}_{mask}$.

\noindent \textbf{Detection Loss.}
Lane detection aims to regress lane coordinates and classification labels. As in~\cite{liao2022maptr}, we employ $L_1$ loss to calculate point-to-point Manhattan distance between the predicted points $\hat{p_{ij}}$ and ground-truth $p_{ij}$. The edge direction loss is also considered in adjacent points by cosine similarity. $\mathcal{L}_{focal}$ calculates the classification loss, which can be formulated as
\begin{small}
\begin{equation}
     \mathcal{L}_{lane}= \sum_{i,j=0}^{M,N}\big(\lambda_{dis}\text{Dis}(\hat{p_{ij}},p_{ij}) +\lambda_{dir}\text{CosSim}
    (\hat{e_{ij}},e_{ij})\big),
\end{equation}
\end{small}
\vspace{-5mm}
\begin{equation}
    \mathcal{L}_{det} = \mathcal{L}_{lane} + \lambda_{cls}\sum_{i=0}^{M}\mathcal{L}_{focal}(\hat{\mathbf{c}_{i}},\mathbf{c}_{i}),
\end{equation}
where $p_{ij}$ is the $j$-th point of the $i$-th instance and $e_{ij}$ is the direction of two adjacent points. $\lambda_{dis}$, $\lambda_{dir}$, and $\lambda_{cls}$ are the weighted factors for the losses of point regression, direction adjustment, and label regression, respectively.

\noindent \textbf{Mask Construction Loss.} Mask learning is able to reduce the risk of overfitting by pixel-level intensive supervision. We use the combination of cross-entropy loss and dice loss~\cite{milletari2016v} to deal with the unbalanced segmentation problems. Given the output masks $\hat{\textbf{M}}_{ins}$ and $\hat{\textbf{M}}_{b}$, we formulate the auxiliary loss as below
\begin{equation}
    \mathcal{L}_{mask} = \lambda_{ins}\mathcal{L}_{ins}(\hat{\textbf{M}}_{ins},\textbf{M}_{ins}) + \lambda_{b}\mathcal{L}_{b}(\hat{\textbf{M}}_{b},\textbf{M}_{b}),
\end{equation}
where $\lambda_{ins}$ and $\lambda_{b}$ are the corresponding loss weights.

\section{Experiments}

\begin{table*}[t]
\begin{center}
\centering
\setlength{\tabcolsep}{9pt}
\resizebox{1\textwidth}{!}{
\begin{tabular}[t]{r||cc||cccc||cccc||c}
\hline\thickhline

\rowcolor{mygray}
&   & &  \multicolumn{4}{c||}{$\text{AP}_{{chamfer}}$} & \multicolumn{4}{c||}{$\text{AP}_{{raster}}$} &  \\
\rowcolor{mygray}
\multicolumn{1}{c||}{\multirow{-2}{*}{Method}} & \multirow{-2}{*}{Backbone} & \multirow{-2}{*}{Epochs}  & ped. & div. & bou. & \textit{avg.} & ped. & div. & bou. & \textit{avg.} & \multirow{-2}{*}{FPS}\\
\hline\hline
\rowcolor{mygray1}
\multicolumn{12}{c}{\textit{Camera-Based Methods}}\\
\hline
HDMapNet~\pub{ICRA22}~\cite{li2022hdmapnet}& EB0 & 30 & 14.4 & 21.7 & 33.0 & 23.0 & - & - & - & - & - \\
InstaGraM~\pub{CVPRW23}~\cite{shin2023instagram}   & EB4 & 30 & 33.8 & 47.2 & 44.0 & 41.7& - & - & - & - & -\\
MapTR~\pub{ICLR23}~\cite{liao2022maptr}   & R50 & 24 & 46.3 & 51.5 & 53.1 & 50.3 & 32.4 &23.5 &17.1 & 24.3  &  15.7 \\
MapTR~\pub{ICLR23}~\cite{liao2022maptr}   & R50 & 30 & 45.2 & 53.8 & 54.3 & 51.1 & 32.9  & 24.9 &  18.9&  25.6  &  15.7 \\
MapVR~\pub{NeurIPS23}~\cite{zhang2023online}   & R50 & 24 & 47.7 &54.4& 51.4& 51.2& 37.5& 33.1 &23.0 &31.2  &15.7 \\
PivotNet~\pub{ICCV23}~\cite{ding2023pivotnet} & R50 & 30 & 53.8 & 55.8 & 59.6 & 57.4 &  -   & -   & -   & -   & 9.6  \\
BeMapNet~\pub{CVPR23}~\cite{qiao2023end}& R50 & 30 & \textbf{57.7} & 62.3 &59.4 &59.8 & - &-& -& - & 4.4  \\

\textbf{MGMap(Ours)}  & R50 & 30 & 57.4 & \textbf{63.5} &  \textbf{63.3} &  \textbf{61.4} &  \textbf{46.5}  &  \textbf{36.5}  &  \textbf{28.7}  & \textbf{37.2} & 11.6 \\
\hdashline
MapTRv2~\pub{arxiv23}~\cite{liao2023maptrv2}& R50 & 24 & 59.8 &62.4 &62.4 &61.5 & - &-& -& - & - \\
\textbf{MGMap*(Ours)}  & R50 & 24 & \textbf{61.8} & \textbf{65.0} &  \textbf{67.5} &  \textbf{64.8} &  -  &  -  &  -  & - & - \\
\hline

VectorMapNet~\pub{ICML23}~\cite{liu2023vectormapnet} & R50 & 110+ft & 42.5 & 51.4 & 44.1 & 46.0 & - & - & - & - & - \\
MapTR~\pub{ICLR23}~\cite{liao2022maptr} & R50 & 110 & 56.2 & 59.8 & 60.1 & 58.7 & 43.6 &  35.7  & 25.8 & 35.0  & 15.7\\
MapVR~\pub{NeurIPS23}~\cite{zhang2023online} & R50 & 110 & 55.0 &61.8 &59.4& 58.8& 46.0& 39.7& 29.9 &38.5   &15.7 \\
BeMapNet~\pub{CVPR23}~\cite{qiao2023end} & R50 & 110 & 62.6  & 66.7  & 65.1  &  64.8 & - & - & - & - & 4.4 \\
\textbf{MGMap(Ours)} & R50 & 110 & \textbf{64.4} & \textbf{67.6} & \textbf{67.7} & \textbf{66.5} & \textbf{54.5} & \textbf{42.1} & \textbf{37.4} & \textbf{44.7} & 11.6 \\ 

\hline
\rowcolor{mygray1}
\multicolumn{12}{c}{\textit{LiDAR-Based Methods}}\\
\hline
HDMapNet~\pub{ICRA22}~\cite{li2022hdmapnet}  & PP  & 30 & 10.4 & 24.1 & 37.9 & 24.1 &- & - & - & - & -\\
VectorMapNet~\pub{ICML23}~\cite{liu2023vectormapnet}  & PP  & 110 & 42.5 & 51.4 & 44.1 & 34.0 &- & - & - & - & -\\

MapTR~\pub{ICLR23}~\cite{liao2022maptr}& Sec  & 24 & 48.5 & 53.7 & 64.7 & 55.6 & 38.9 &30.1 &41.1 & 36.7  &  6.0 \\
  
\textbf{MGMap(Ours)}  & Sec  & 24 & \textbf{63.5} &  \textbf{66.7} & \textbf{73.6} & \textbf{67.9} & \textbf{53.4} &\textbf{44.4} &\textbf{52.6} & \textbf{50.1}& 5.5 \\ 
\hline
\rowcolor{mygray1}
\multicolumn{12}{c}{\textit{Camera-LiDAR Fusion Methods}}\\
\hline
HDMapNet~\pub{ICRA22}~\cite{li2022hdmapnet} & EB0 \& PP & 30 & 16.3 & 29.6 & 46.7 & 31.0 & - & - & - & - & - \\
VectorMapNet~\pub{ICML23}~\cite{liu2023vectormapnet}  & EB0 \& PP & 110+ft & 48.2 & 60.1 & 53.0 & 53.7 & - & - & - & - & - \\

MapTR~\pub{ICLR23}~\cite{liao2022maptr} & R50 \& Sec & 24 & 55.9 & 62.3 & 69.3 & 62.5 &46.4&38.4&49.2&44.7& 5.2 \\
MapVR~\pub{NeurIPS23}~\cite{zhang2023online} & R50 \& Sec & 24 &60.4& 62.7 &67.2& 63.5 &52.4& 46.4& 54.4 &51.1  &5.2 \\
\textbf{MGMap(Ours)} & R50 \& Sec & 24 & \textbf{67.7} & \textbf{71.1} & \textbf{76.2} & \textbf{71.7} & \textbf{59.6} & \textbf{47.3} & \textbf{54.6}& \textbf{53.8} & 4.8\\ 
\hline\thickhline
\end{tabular}}
\end{center}
\vspace{-5mm}
\caption{Quantitative evaluation of map vectorization on \textit{nuScenes} \textit{val}. at $60m\times30m$ perception range under different input modalities and backbone settings, 
``EB0", ``EB4", ``R50", and ``Sec" correspond to the backbones Efficient-B0, Efficient-B4~\cite{tan2019efficientnet}, ResNet50~\cite{he2016deep}, and SECOND~\cite{yan2018second} for LiDAR, respectively. ``ft" means the two-stage fine-tune strategy. ``MGMap*" means the reimplemented structure based on stronger MapTRV2~\cite{qiao2023end}. The inference speed is measured on the same computer with a single NVIDIA Tesla V100 GPU. 
}
\vspace{-4mm}
\label{tab:exp_nus}
\end{table*}

\subsection{Datasets and Benchmarks}
We conduct extensive experiments on two public datasets, including \textit{nuScenes}~\cite{nuScenes} and \textit{Argoverse2}~\cite{Argoverse2}. \textit{nuScenes} dataset contains 1000 driving scenes collected from Boston and Singapore. 750 and 150 scene sequences are annotated for training and validation, respectively.  
Each scene sequence consists of 40-keyframe data with a sampling rate of 2Hz. 
For each keyframe, there are 6 PV images and the corresponding point clouds from 32-beam LiDAR. \textit{Argoverse2} contains 1000 scenes collected from six cities with 7 PV images. We use a subset of \textit{Argoverse2}, which is provided by the Online HD Map Construction Challenge\footnote{https://github.com/Tsinghua-MARS-Lab/Online-HD-Map-Construction-CVPR2023}. 
We mainly focus on three map elements, including lane-divider (div.), ped-crossing (ped.), and road boundary (bou.).

\begin{figure*}[t]
  \includegraphics[width=1\textwidth]{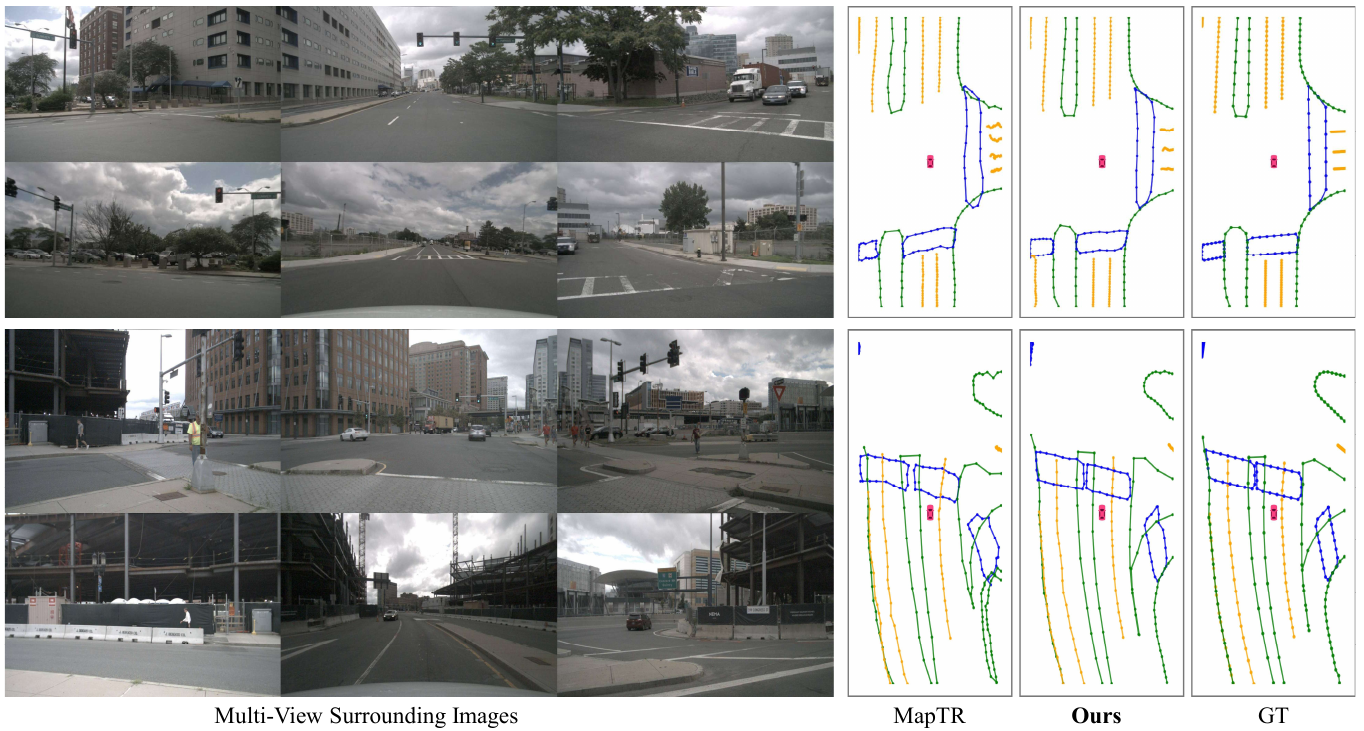}
  \vspace{-6mm}
  \caption{The visual results of MapTR~\cite{liao2022maptr}, our proposed \textbf{MGMap} approach and the corresponding ground truth.}
  \label{fig5}
  \vspace{-3mm}
\end{figure*}

\subsection{Evaluation Metrics}
To facilitate comprehensive evaluations, we employ the Chamfer distance-based metrics, including average precision $\text{AP}_{{chamfer}}$~\cite{li2022hdmapnet} and the IoU-based average precision $\text{AP}_{raster}$~\cite{zhang2023online}, which evaluates the model from the point-coordinate aspect considers each map element as a whole unit separately. This ensures that the map vectorization quality can be assessed from different perspectives.  

\noindent \textbf{Chamfer-Distance AP.} For fair comparisons, $\text{AP}_{chamfer}$ is inherited from previous map vectorization works~\cite{liao2022maptr,liu2023vectormapnet}. Specifically, the average Euclidean distances between sampled points in each map line and the nearest points in the ground-truth labels are measured. The AP is calculated under the average of three distance thresholds, $\tau_{chamfer} \in \{0.5m, 1m, 1.5m\}$. Each prediction is treated as true-positive (TP) when the distance is below the threshold. $\text{AP}_{chamfer}$ evaluates map construction quality from point-level perspectives, since it calculates the distances between points as the measured errors.

\noindent \textbf{IoU-based AP.} Following MapVR~\cite{zhang2023online}, $\text{AP}_{raster}$ is calculated by the Intersection over Union (IoU) among pixels. Both predictions and ground truth labels are rasterized into polylines in HD maps. Raster map size is set to $480 \times 240$, and each map element is dilated by two pixels on each side. Detection quality is evaluated by calculating the IoU of the rasterized representations between the prediction and ground truth. The calculation of AP is set under thresholds $\tau_{IoU} \in \{0.25:0.5:0.05\}$ for line-shaped elements and $\tau_{IoU} \in \{0.5:0.75:0.05\}$ for polygon-shaped pedestrian crossings, where 0.05 represents the step size for the change of threshold. $\text{AP}_{raster}$ is calculated under the average of all the thresholds among the upper and lower bounds. Thus, $\text{AP}_{raster}$ considers each lane line at the pixel level and evaluates map quality from the whole instance perspective.

\subsection{Implementation Details}
To ensure fair comparisons, we choose ResNet50 \cite{he2016deep} as the image backbone. SECOND~\cite{yan2018second} is adopted as the backbone for LiDAR modality. The BEV size, defined as $H_{BEV} \times W_{BEV}$, is set to $200 \times 100$. The maximum number of instances and point queries are set to 50 and 20, respectively. In PG-MPR module, the hyper-parameter $d$ is set to 0.1, which is the normalized patch size. 
We use AdamW~\cite{reddi2019convergence} optimizer with a learning rate of $6e^{-4}$. All models are trained on 8 NVIDIA Tesla V100 GPUs with a batch size of 6 per GPU. The weighted factors for detection loss are the same as those in MapTR~\cite{liao2022maptr}. For mask branches, $\{\lambda_{ins}, \lambda_{b}\}$ are set to $\{2, 15\}$, respectively.

\subsection{Main Results}
\noindent \textbf{Performance on \textit{nuScenes} Dataset.} 
As shown in \Cref{tab:exp_nus}, we compare MGMap approach against the SOTA methods on the validation set of \textit{nuScenes} with different settings. The experimental performance is evaluated under $\text{AP}_{chamfer}$ and $\text{AP}_{raster}$. 
It can be seen that our proposed approach outperforms previous methods and obtains the best performance. 
Compared with baseline MapTR~\cite{liao2022maptr}, MGMap archives 10.3 mAP improvement with multi-view camera-based input under the same settings of ResNet-50 and 30 epochs for training. 
It is worth noting that MGMap achieves 67.9 mAP for LiDAR and 71.7 mAP for fusing camera data with LiDAR, which demonstrates the strong generalization capability of our scheme. Furthermore, the visual results of MGMap in several driving scenarios are shown in~\Cref{fig5}. More comparison results under different conditions are included in the supplementary material.
\begin{table}[t]
\begin{center}
\footnotesize
\centering
\renewcommand\tabcolsep{4.7pt}
\renewcommand\arraystretch{1.15}
\begin{tabular}{c||c||c cccc}
            \hline\thickhline
			\rowcolor{mygray}
			Method & Backbone & AP$_{\textit{ped.}}$ & AP$_{\textit{div.}}$ & AP$_{\textit{bou.}}$ &  mAP   \\
            \hline\hline
			HDMapNet~\cite{li2022hdmapnet} & EB0 & 5.7 & 13.1 & 37.6 &  18.8 \\
			VectorMapNet~\cite{liu2023vectormapnet} & R50 & 37.2 & 50.4 & 40.6 & 42.7 \\
			MapTR~\cite{liao2022maptr} & R50 & 50.6 & 60.7 & 61.2 & 57.4 \\
			\hline
         
			  MGMap(Ours) & R50 &  \textbf{52.8} & \textbf{67.5} & \textbf{68.1} & \textbf{62.8} \\
			\hline\thickhline
		\end{tabular}

\end{center}
\vspace{-6.0mm}
\caption{Performance comparison with baseline methods at $60m\times30m$ perception range on a subset of \textit{Argoverse2} provided by the Online HD Map Construction Challenge.}
\vspace{-6.0mm}
\label{tab:exp_argo}
\end{table}

\noindent \textbf{Performance on \textit{Argoverse2} Dataset.} 
By following the settings on the online HDMap construction challenge, we re-implement MapTR and MGMap on the \textit{Argoverse2} dataset. \Cref{tab:exp_argo} presents our experimental results\textbf{}. It can be observed that our method achieves competitive performance on the \textit{Argoverse2} dataset, MGMap achieves 5.4 mAP improvement compared with MapTR, which further shows the effectiveness of our proposed approach.

\noindent \textbf{Performance on Enlarged Perception Ranges.}
To evaluate the robustness of model, we conduct experiments on the enlarged perception ranges. Under the same settings, we re-implement MapTR and our MGMap with the perception ranges of $60m\times60m$ and $30m\times90m$ at the X-axis and Y-axis in BEV space, in which query numbers are proportionally enlarged to account for the basic property. All models are trained for 30 epochs. \Cref{tab:exp_long} reports the experiment results. Compared to MapTR, our MGMap achieves consistent performance improvements, with a 9.5 mAP improvement on the $60m\times60m$ perceptions range setting and a 10.2 mAP improvement on the $30m\times90m$ range setting.

\begin{table}[ht]
\begin{center}
\footnotesize
\centering
\renewcommand\tabcolsep{5pt}
\renewcommand\arraystretch{1.15}

\begin{tabular}[t]{c c|| cccc}

\hline\thickhline
\rowcolor{mygray}
BEV range & Method & AP$_{\textit{ped.}}$ & AP$_{\textit{div.}}$ & AP$_{\textit{bou.}}$ &  mAP \\

\hline\hline

\multirow{3}{*}{$60m\times60m$} & MapTR~\cite{liao2022maptr} & 37.9  & 44.4  &  41.6 & 41.3  \\

   & MGMap(Ours) & 48.7 & 53.4 & 50.0 &  50.8 \\
   & $\Delta$mAP &  \textbf{+10.8}  & \textbf{+9.0}  & \textbf{+8.4} & \textbf{+9.5}    \\
\hline
\multirow{3}{*}{$90m\times30m$} & MapTR~\cite{liao2022maptr} & 36.5 &  43.7 & 39.1 & 39.8  \\

   & MGMap(Ours) & 46.4 &  56.1 &  49.1 & 50.5   \\
   & $\Delta$mAP & \textbf{+9.9}  & \textbf{+12.4}  & \textbf{+10.0} & \textbf{+10.2}   \\
\hline\thickhline
\end{tabular}
\end{center}
\vspace{-6mm}
\caption{Experimental results with enlarged perception range settings on the \textit{nuScenes} dataset. Our proposed MGMap approach outperforms MapTR significantly on all evaluation metrics.}
\label{tab:exp_long}
\vspace{-5mm}
\end{table}

\subsection{Ablation Study}
In this section, ablation experiments are conducted to examine the effectiveness of our proposed modules and designs. For fair comparisons, all experiments are conducted on the camera modality of the \textit{nuScenes} dataset with 30 epochs for training. ResNet50 is employed as the image backbone. 

\noindent \textbf{Ablation on Mask-Guided Design.} Our mask-guided design comprises two main components: the MAI decoder and the PG-MPR module, which are utilized to handle the overall shape structure and specific details from the instance level and the point level. The MAI construction leverages the masks to obtain global structure information, which contributes to shape understanding. Furthermore, the patch refinement considers the local level more precisely. This strategy allows for more specific and fine-grained adjustments to individual points, which contributes to precise localization. The ablation results presented in \Cref{tab:ablation_inspoint} demonstrate the impact of each level design. When compared to the absence of mask guidance, the inclusion of the MAI decoder and PG-MPR module leads to individual improvements of 1.9 mAP and 2.6 mAP, respectively. Furthermore, the combination of these two components in the overall design achieves the highest performance, with a mAP of 61.4.

\noindent \textbf{Ablation on EML Neck.} To investigate the effectiveness of the EML neck design, we conduct ablation experiments to compare it with FPN in PV image space. Multi-scale BEV features contain enriched semantic and location information with larger receptive fields, which is beneficial for the positioning of irregular shapes. As shown in \Cref{abl1}, compared to baselines without multi-level features, EML design at BEV space (referred to as ``BEV'') achieves a larger performance gain. While PV space design fails to achieve the expected effect (``PV''). Besides, enhanced BEV features also contribute to mask generation and boost the performance of the mask-guided design (``MG'').

\noindent \textbf{Ablation on PG-MPR Design.} Finally, we investigate the ablation experiment on the setting of the position-guided mask patch refinement module. In our experiment settings, we compare different selections of patch size $d$ and the number of refinement stages $s$. 
The evaluated normalized patch sizes range from 0.08 to 0.12, and refinement stages are set from 1 to 3. As illustrated in~\Cref{rego_ab}, applying the refinement strategy contributes to performance improvement. 

\begin{table}[ht!]
\vspace{-2mm}
\footnotesize
  \centering
  \renewcommand\arraystretch{1.15}
  \renewcommand\tabcolsep{10pt}
  \begin{tabular}{ cc||cc c c}
    \hline\thickhline
    \rowcolor{mygray}
    Ins. & Point& AP$_{\textit{ped.}}$ & AP$_{\textit{div.}}$ & AP$_{\textit{bou.}}$ &  mAP \\ 
    \hline\hline
     &   & 53.1 & 60.1 & 59.5 & 57.6    \\
     \cmark &   & 55.4 & 61.7& 61.5 &  59.5     \\ 
     & \cmark  & 54.9  & 63.4 & 62.3 &60.2      \\
     \cmark & \cmark  & \textbf{57.4}  & \textbf{63.5} & \textbf{63.3} & \textbf{61.4}    \\
    \hline\thickhline
    \end{tabular}
  \vspace{-2mm}
  \caption{Ablation study of mask-guided design at the instance and point levels. Ins. denotes the instance-level MAI decoder and Point means the point-level PG-MPR module design.}
  \label{tab:ablation_inspoint}
 \vspace{-3mm} 
\end{table}
\begin{table}[ht!]

\footnotesize
\centering
\renewcommand\tabcolsep{4pt}
\begin{center}

\label{tab:ab_multiscale}
\renewcommand\tabcolsep{5.2pt}
 \renewcommand\arraystretch{1.15}
 \begin{tabular}{cccc||cc cc}
    \hline\thickhline
    \rowcolor{mygray}
    Baseline&PV & BEV& MG&AP$_{\textit{ped.}}$ & AP$_{\textit{div.}}$ & AP$_{\textit{bou.}}$ &  mAP \\ 
    \hline\hline

     \cmark&&   &   & 45.2 & 53.8 &  54.3& 51.1   \\
     \cmark&\cmark &  &  &  41.7& 49.9 &52.6  & 48.1    \\ 
    \cmark&\cmark & & \cmark & 47.3 & 54.1 & 55.1 & 52.2   \\
      \cmark&  &\cmark  &  & 53.1 & 60.1& 59.5 & 57.6   \\
   
    \cmark& &\cmark  & \cmark & \textbf{57.4} & \textbf{63.5} & \textbf{63.3} & \textbf{61.4}  \\
    \hline\thickhline
    \end{tabular}
\vspace{-2mm}
\caption{Ablation study of EML neck design by investigating the performance of multi-level features at PV and BEV stages. Empirical results show that the best performance is achieved by taking advantage of the BEV-level EML neck for mask-guided design.}\label{abl1}
\vspace{-6mm}
\end{center}
\end{table}

\begin{table}[ht!]
\footnotesize
\centering
\renewcommand\tabcolsep{10.7pt}
\renewcommand\arraystretch{1.15}
  \begin{tabular}{cc||cc c c}
    \hline\thickhline
    \rowcolor{mygray}
    $d$ & $ s $& AP$_{\textit{ped.}}$ & AP$_{\textit{div.}}$ & AP$_{\textit{bou.}}$ &  mAP \\ 
    \hline\hline
    $/$ & $/$&55.4 &61.7 & 61.5 & 59.5 \\
    0.08& 2 & 55.8  & 64.3  & 62.3 & 60.8     \\
     0.1& 1 & 56.3  & 62.6 & 62.5  &   60.5   \\
    \textbf{0.1}& \textbf{2} & \textbf{57.4} & \textbf{63.5}  & \textbf{63.3} & \textbf{61.4}   \\
    0.1& 3 & 56.0  & 62.9  & 62.9 &  60.6   \\
   0.12& 2  & 56.0 &62.7&  61.5 &  60.1    \\
    \hline\thickhline
    \end{tabular}
  \vspace{-2mm}
  \caption{Performance of point-level PG-MPR design with different patch sizes ($d$) and refinement stages ($s$). The first row means the result without the point-level refinement.}
  \label{rego_ab}
  \vspace{-6mm}
\end{table}

\noindent Experiments also indicate that the oversized patches will introduce irrelevant information and adversely affect performance while undersized patches lead to information loss and also result in suboptimal results. The best performance is achieved when the selected patch size is set to 0.1 along with the two-stage refinement.  
\section{Conclusion}
In this paper, we propose MGMap, an effective approach to online HD map vectorization with the guidance of learned masks. By taking advantage of masks at both the instance and point levels, we alleviate the challenges of rough detection and loss of details arising from subtle and sparse annotations in HD maps. Our proposed MGMap not only demonstrates state-of-the-art performance but also exhibits strong robustness in online map vectorization across various experimental settings. For future work, fusing with other perceptual tasks to construct a more comprehensive representation is still a direction worth exploring, which holds promise for further advancements in autonomous driving.

\section*{Acknowledgments}
This work is supported by National Natural Science Foundation of China under Grants (62376244). It is also supported by Information Technology Center and State Key Lab of CAD\&CG, Zhejiang University.
\clearpage
\appendix
\maketitlesupplementary

\setcounter{figure}{0}
\setcounter{table}{0}
\renewcommand{\thefigure}{A\arabic{figure}}
\renewcommand{\thetable}{A\arabic{table}}

\noindent In this supplementary file, we provide more details, discussions, and experiments as follows:
\begin{itemize}
    \item More details of our method;
    \item Additional experiments;
    \item Extensive qualitative results;
    \item Limitations and future work.
\end{itemize}

\section{More Details of Our Method}
\label{details}
\subsection{Enhanced Multi-level Neck}
In this section, we present more details of the fused attention in the enhanced multi-level (EML) neck. 
EML neck consists of three cascaded layers, each layer is a basic ResNet block~\cite{he2016deep} with fused channel attention (CA) and spatial attention (SA)~\cite{woo2018cbam}, which focus on semantics and positional information to adaptively learn the crucial regions of the BEV space. 

For the BEV feature with the shape of $D \times H_{BEV}\times W_{BEV}$, channel attention calculates channel weight of size $D\times 1\times 1$ through average pooling and max pooling, 
emphasizing diverse channel information.
A sigmoid function after MLP is employed to calculate the channel-wise attention map. For spatial attention, average pooling and max pooling are used to compress the channel dimension $D$, in which we construct spatial weight with shape $1\times H_{BEV}\times W_{BEV}$ to obtain the weight of location information. Then, a spatial map can be generated by a $7\times 7$ convolutional layer followed by the sigmoid function.

\subsection{Auxiliary Loss Setting}
In addition to regressing the point's position, an auxiliary loss for mask construction is required. 
As mentioned in the main paper, we combine cross-entropy loss $\mathcal{L}_{ce}$ and dice loss $\mathcal{L}_{dice}$~\cite{milletari2016v} to construct instance mask $\hat{\textbf{M}}_{ins}$ and binary mask $\hat{\textbf{M}}_b$. Specifically, 
\begin{small}
\begin{equation}
    \mathcal{L}_{ins} = \lambda_{ins}\mathcal{L}_{ce}(\hat{\textbf{M}}_{ins},\textbf{M}_{ins}) + \lambda_{d1}\mathcal{L}_{dice}(\hat{\textbf{M}}_{ins},\textbf{M}_{ins}),
\end{equation}
\vspace{-2mm}
\begin{equation}
    \mathcal{L}_{b} = \lambda_{b}\mathcal{L}_{ce}(\hat{\textbf{M}}_{b},\textbf{M}_{b}) + \lambda_{d2}\mathcal{L}_{dice}(\hat{\textbf{M}}_{b},\textbf{M}_{b}),
\end{equation}
\end{small}
where $\{\lambda_{ins},\lambda_{d1},\lambda_{b},\lambda_{d2}\}$ are the corresponding loss weights for two level masks. 
$\mathcal{L}_{ce}$ calculates the loss of each pixel equally, while $\mathcal{L}_{dice}$ takes consideration of mining the foreground areas, which can be formulated as below
\begin{equation}
    \mathcal{L}_{dice} = 1-2\cdot\frac{\hat{\textbf{M}}\cap\textbf{M}}{\hat{\textbf{M}}\cup\textbf{M}}. 
\end{equation}
To this end, we expect a larger intersection over the union area for the predicted mask $\hat{\textbf{M}}$ and ground truth $\textbf{M}$.

\section{Additional Experiments}
\label{exp}
\subsection{Time Analysis}
\Cref{time_analy} shows the detailed time analysis of each component. Compared with MapTR~\cite{liao2022maptr}, EML neck and PG-MPR bring a slight time delay while the initial BEV extraction causes the main time-consuming. 
\vspace{-2mm}
\begin{table}[ht]
    \centering
    \footnotesize
    \renewcommand\tabcolsep{3.5pt}
    \renewcommand\arraystretch{1.2}
 \begin{tabular}{c||cc|c|c|c}
      \hline\thickhline
      
    \multirow{2}{*}{Method} &\multicolumn{2}{c|}{BEV Extractor} & \multirow{2}{*}{decoder}& \multirow{2}{*}{PG-MPR}&\multirow{2}{*}{Total}\\
    
     &init.BEV&neck& & & \\

     \hline\hline
    MapTR&48.4ms&/&16.3ms&/&64.7ms\\
    \hline
    Ours&48.4ms&6.9ms&16.3ms&12.7ms&84.3ms\\
      \hline\thickhline
    \end{tabular}
 \vspace{-2mm}
    \caption{ Detailed runtime analysis and comparison with MapTR}
     \vspace{-6mm}
    \label{time_analy}
\end{table}
    
\subsection{Experiments under Different Conditions}
We compare MGMap with the state-of-the-art method MapTR~\cite{liao2022maptr} under different weather and lighting conditions, in which the nuScenes dataset~\cite{nuScenes} is split by~\cite{qiao2023end}. We employ ResNet-50~\cite{he2016deep} as the image backbone and SECOND~\cite{yan2018second} for the LiDAR modality. All models are trained for 24 epochs. Moreover, experiments are conducted under $\text{mAP}_{chamfer}$ with a threshold setup of [0.5m, 1.0m, 1.5m]. As illustrated in \Cref{weather}, our method achieves the stable improvements with more than +9 mAP under different conditions.
\begin{table}[ht]
    \centering
    \footnotesize
    \renewcommand\tabcolsep{3.5pt}
    \renewcommand\arraystretch{1.5}
    \begin{tabular}{cc||cccccc}
    \hline\thickhline
      Modality  &  Method &sunny&cloudy&rainy&day&night&mAP\\
    \hline\hline
        \multirow{2}{*}{Camera}&MapTR& 53.5&49.7 & 43.3&50.5&35.7&50.0 \\
                  &MGMap&\textbf{65.2}   &\textbf{62.8}&\textbf{49.4} &\textbf{61.7}&\textbf{37.6}& \textbf{60.8}\\
    \hline
        \multirow{2}{*}{LiDAR}&MapTR&  59.2 &56.9 & 46.7&56.3&38.9&55.8 \\
         &MGMap&\textbf{71.6} & \textbf{70.6}& \textbf{54.0}&\textbf{68.0}&\textbf{48.8}&\textbf{67.5} \\
    \hline
        \multirow{2}{*}{Fusion}&MapTR& 66.6&62.7 &54.5 &63.4&44.8& 62.8\\
         &MGMap& \textbf{74.7}  &\textbf{75.9} &\textbf{59.5} &\textbf{72.2}&\textbf{53.3}&\textbf{ 71.7}\\
    \hline\thickhline
    \end{tabular}
    \vspace{-2mm}
    \caption{Comparisons under several weather and lighting conditions with different input modalities, our MGMap approach consistently achieves significant improvements over MapTR.}
    \label{weather}
\end{table}
\vspace{-5mm}

\subsection{Ablations on Auxiliary Loss}
In this section, we investigate the effects of auxiliary losses. As mentioned in the main paper, we present a parallel branch for mask predictions, which requires intensive supervision at the BEV space to construct the feature-prominent masks. Table~\ref{aux_loss} reports the experimental results. 
Compared with the model without auxiliary loss, mask construction introduces intensive pixel-level learning and alleviates the overfitting issue to some extent, resulting in a noteworthy +1.7 mAP improvement (57.6  \textit{v.s.} 59.3).
However, simple parallel segmentation learning lacks full use of mask features. To synergize with the vectorization task, our mask-guided design is proposed to boost the potential of mask features and obtains the best performance with 61.4 mAP.

\begin{table}[ht]
    \centering
    \footnotesize
    \renewcommand\tabcolsep{5pt}
    \renewcommand\arraystretch{1.25}
    \begin{tabular}{r||cccc}
     \hline\thickhline
      \multicolumn{1}{c||}{Strategy} & AP$_{ped.}$&AP$_{div.}$&AP$_{bou.}$&mAP \\
       \hline\hline
     w/o. mask &53.1&60.0&59.5&57.6 \\
     +parallel segementation&53.3&63.0&61.6&59.3\\
     +mask-guided design & \textbf{57.4} & \textbf{63.5} & \textbf{63.3} & \textbf{61.4} \\
    \hline\thickhline
    \end{tabular}
    \vspace{-2mm}
    \caption{Ablation studies on auxiliary loss. Adding parallel segmentation achieves a certain level of improvement, while mask-guided design further enhances performance with the best result.}
    \vspace{-6mm}
    \label{aux_loss}
\end{table}
\begin{figure*}[th]
\centering
  \includegraphics[width=1\textwidth]{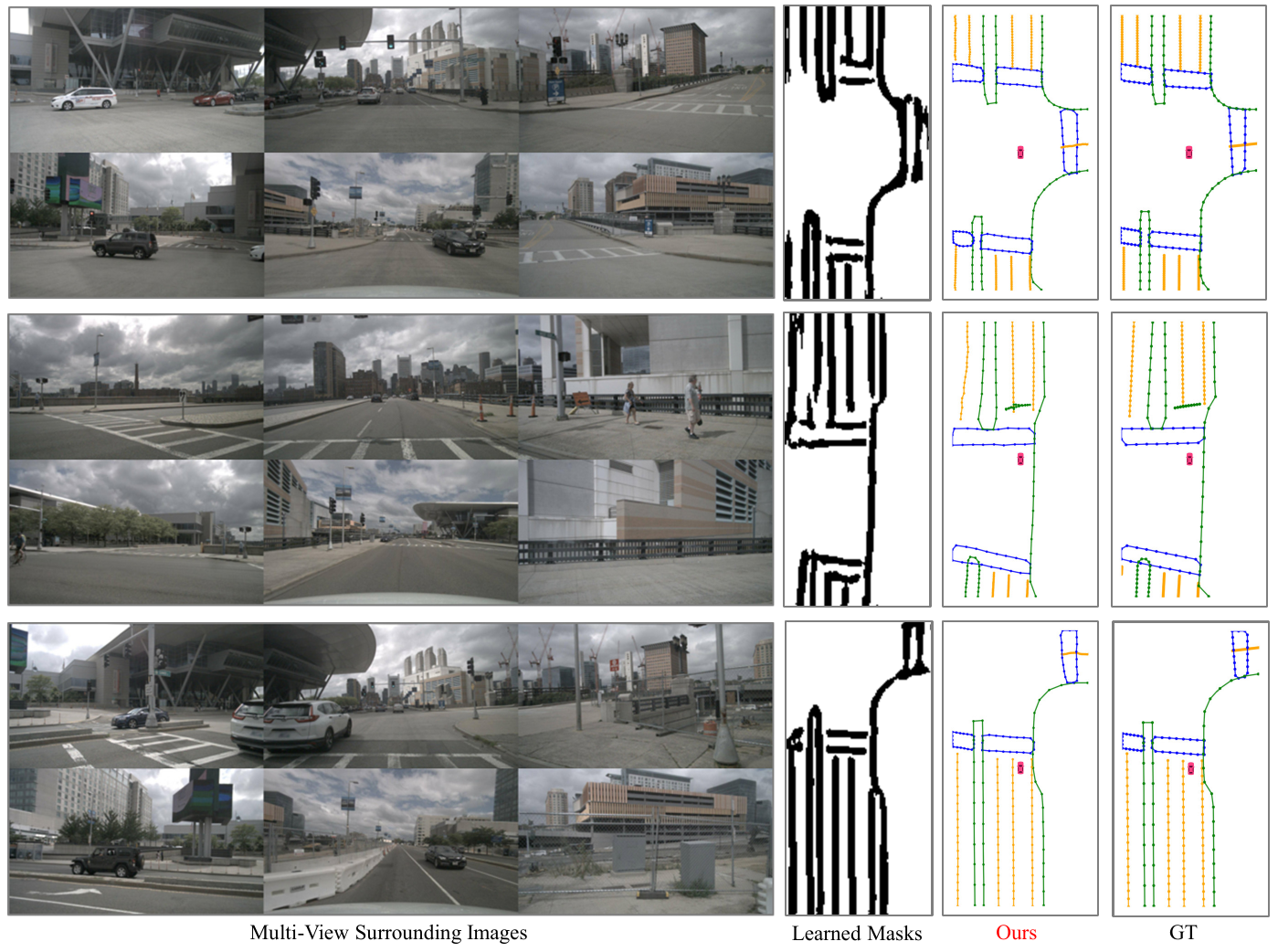}
  \vspace{-6mm}
  \caption{The visual results of the learned masks, our proposed \textbf{MGMap} approach and the corresponding ground truth.}
  \label{fig_best}
  \vspace{-2mm}
\end{figure*}
\section{Extensive Qualitative Results}
\label{vis}
\begin{center}
    \begin{figure*}[ht]
    \centering
  \includegraphics[width=0.8\textwidth]{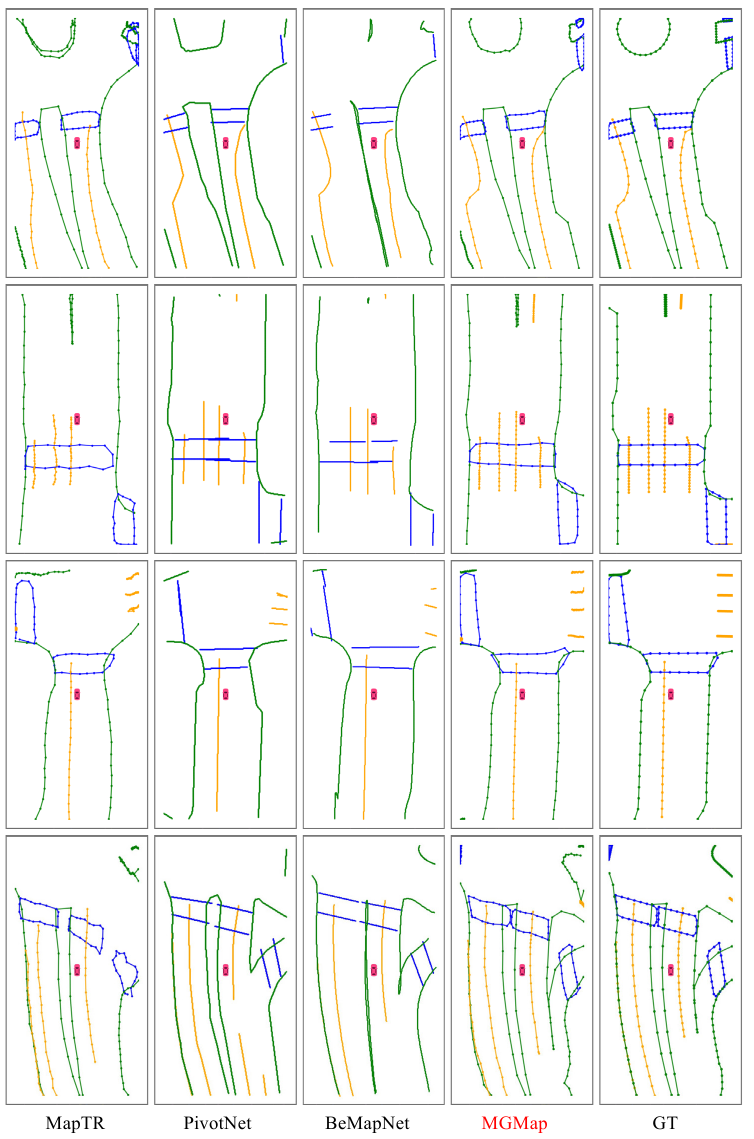}
  \vspace{-2mm}
  \caption{Comparison with recent sota methods.}
  \label{fig_comp}
  \end{figure*}
\end{center}

\begin{center}
    \begin{figure*}[ht]
    \centering
  \includegraphics[width=1.0\textwidth]{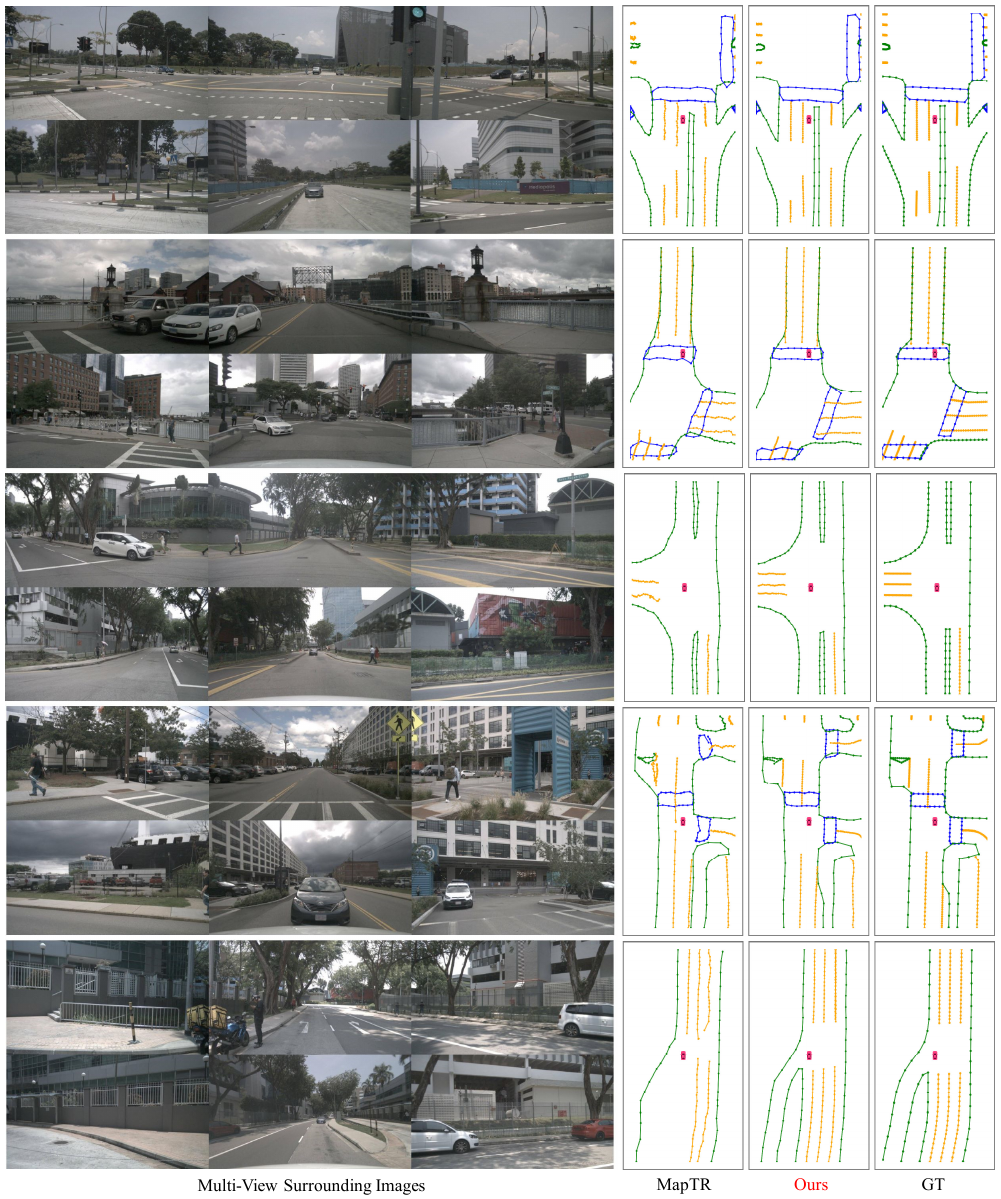}
  \vspace{-5mm}
  \caption{Visualization results under the weather condition of \textit{sunny}.}
  \label{fig_sunny}

\end{figure*}
\end{center}

\begin{center}
    \begin{figure*}[ht]
  \includegraphics[width=1.0\textwidth]{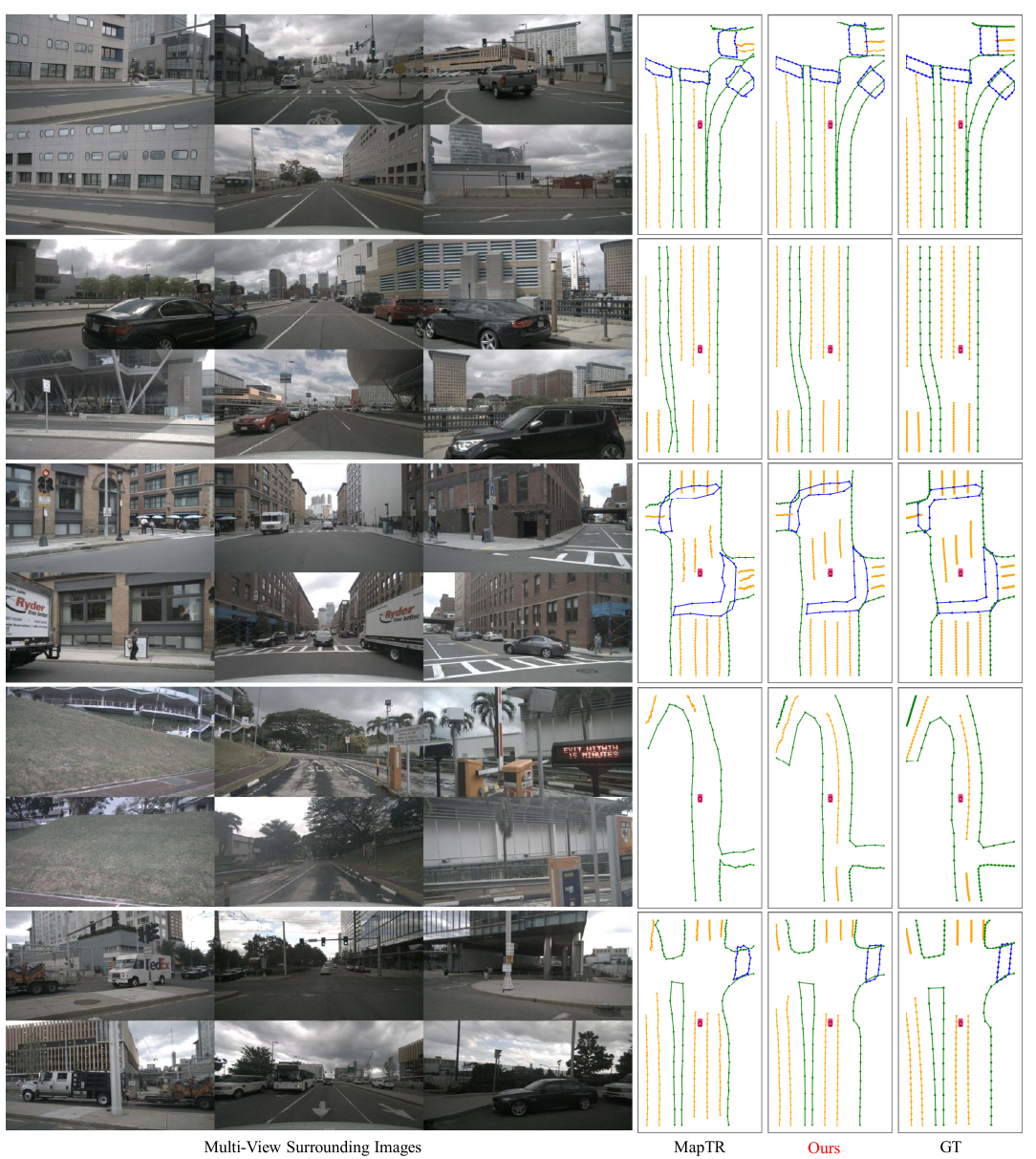}
  \vspace{-5mm}
  \caption{Visualization results under the weather condition of \textit{cloudy}.}
  \label{fig_cloudy}
\vspace{4mm}
\end{figure*}
\end{center}

\begin{center}
    \begin{figure*}[ht]
  \includegraphics[width=1.0\textwidth]{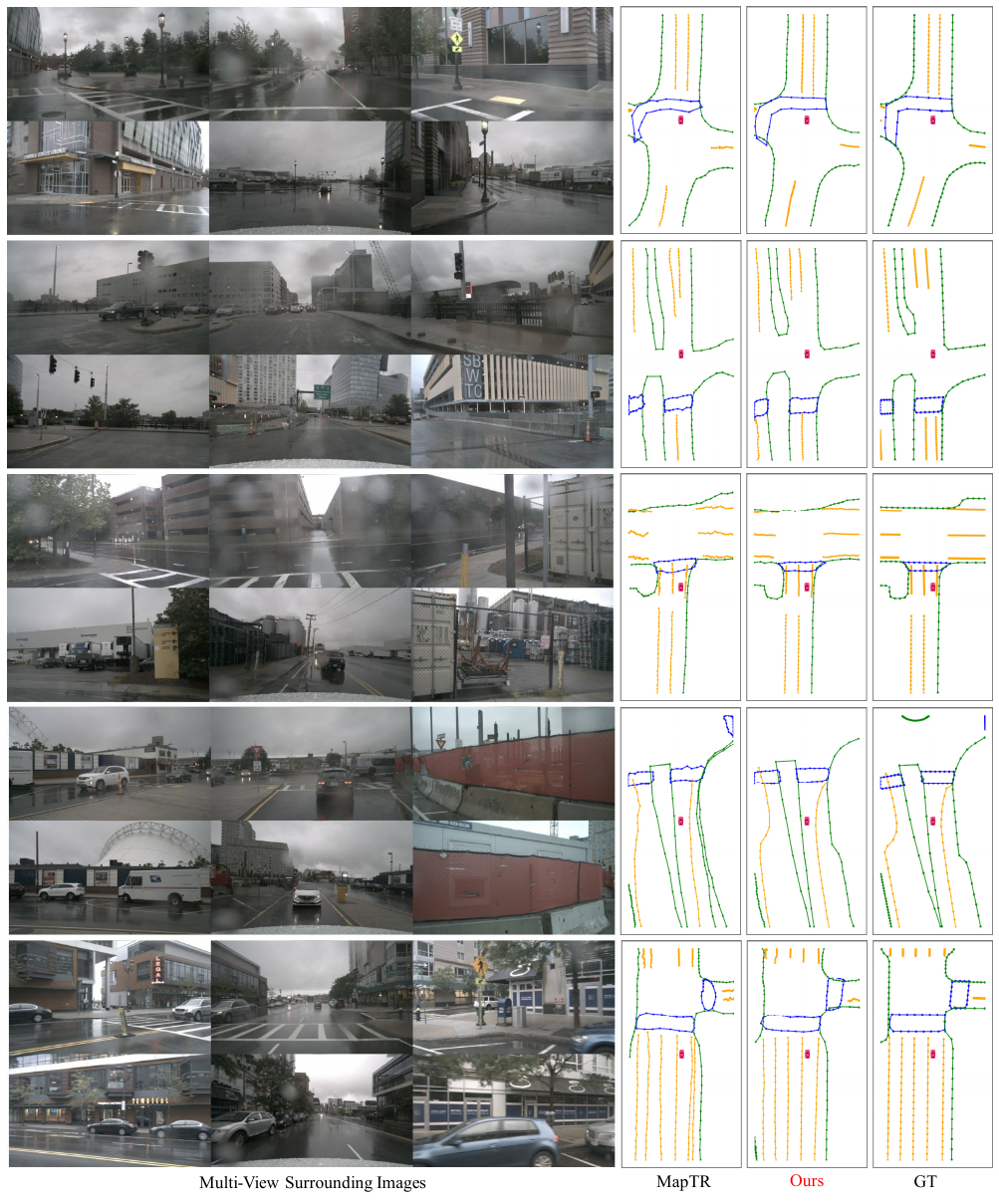}
  \vspace{-7mm}
  \caption{Visualization results under the weather condition of \textit{rainy}.}
  \label{fig_rainy}

\end{figure*}
\end{center}

\Cref{fig_best} presents the visualization results of the learned masks and the final predictions, in which the binary masks are constructed to assist for the final map vectorization. ~\Cref{fig_comp} provides the visual comparison with recent sotas. Later, ~\Cref{fig_sunny,fig_cloudy,fig_rainy,fig_night} provide extensive visualization results of our MGMap, comparing with the state-of-the-art approach MapTR under different weather and lighting conditions. Our MGMap method consistently demonstrates its promising capabilities across various scenarios.

    \section{Limitations and Future Work}
\begin{center}
    \begin{figure*}[ht]
  \includegraphics[width=1.0\textwidth]{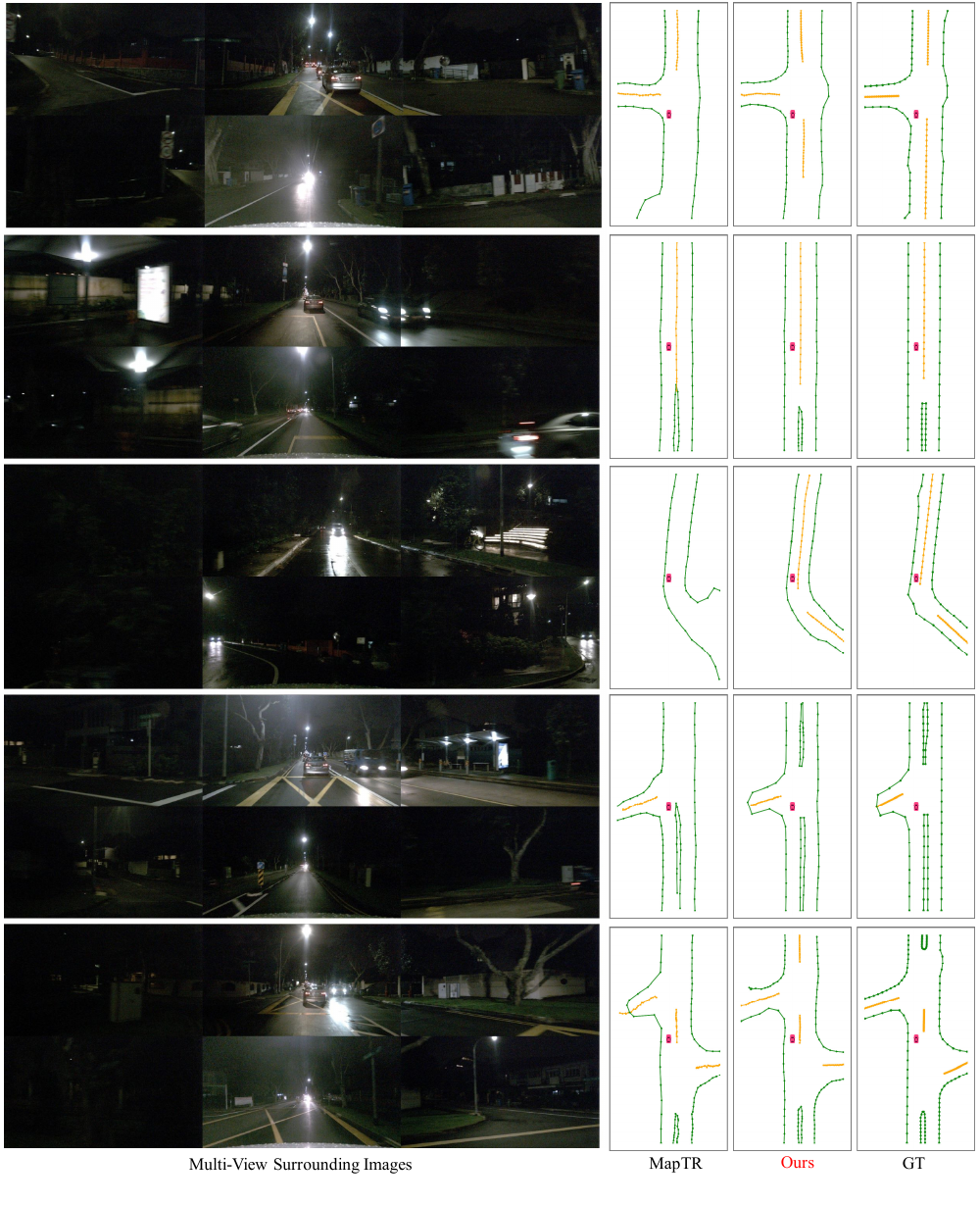}
  \vspace{-10mm}
  \caption{Visualization results under the \textit{night} condition.}
  \label{fig_night}

\end{figure*}
\end{center}

As shown in \Cref{fig_night}, under some adverse conditions, like low light, occlusion, and long-range perceptions, our image-based approach still has limitations in achieving reliable performance. It is mainly caused by the lack of effective features and the inferior interpretation of driving scenes. In the future, multi-modal fusion, temporal information, and the introduction of road priors will be explored to address the current shortcomings and obtain the vectorized HD map with higher precision.

\vspace{30mm}

\clearpage
{
    \small
    \bibliographystyle{ieeenat_fullname}

}
\end{document}